\documentclass[letterpaper, 10 pt, conference]{ieeeconf}  
\IEEEoverridecommandlockouts                              
\usepackage{graphics} 
\usepackage{amsmath} 
\usepackage{amssymb}  
\usepackage[utf8]{inputenc}
\usepackage[OT1]{fontenc}
\usepackage[nospace]{cite}
\usepackage{fancyhdr}

\usepackage{textcomp}\usepackage{gensymb}

\usepackage{multicol}

\usepackage{subfig}
\usepackage{graphicx}

\usepackage{xr}
\usepackage{xr-hyper}

\usepackage[nolist]{acronym}
\usepackage{nicefrac}
\usepackage{upgreek}
\usepackage{units}
\usepackage{afterpage}
\usepackage{booktabs}
\usepackage[binary-units]{siunitx}
\usepackage{multicol}
\usepackage[bookmarks=true]{hyperref}
\usepackage{mathdots} 
\usepackage{tablefootnote}
\usepackage[final]{microtype}

\newcommand{\norm}[1]{\left\lVert#1\right\rVert}
\newcommand{\mat}[1]{{\ensuremath{{\mathbf{#1}}}}}

\newcommand\footnoteref[1]{\protected@xdef\@thefnmark{\ref{#1}}\@footnotemark}

\DeclareSIUnit[per-mode=symbol,per-symbol=p]{\MBps}{\mega\byte\per\second}
\DeclareSIUnit[per-mode=symbol,per-symbol=p]{\Mbps}{\mega\bit\per\second}
\DeclareSIUnit[per-mode=symbol,per-symbol=p]{\MB}{\mega\byte}
\title{\LARGE \bf
Collaborative Robot Mapping using Spectral Graph Analysis
}
\author{Lukas Bernreiter$^{1}$, Shehryar Khattak$^{2}$, Lionel Ott$^{1}$, Roland Siegwart$^{1}$, Marco Hutter$^{2}$ and Cesar Cadena$^{1}$
\thanks{This  work  was  supported  by  the  National  Center  of  Competence  in Research (NCCR) Robotics through the Swiss National Science Foundation, and the European Union’s Horizon 2020 research and innovation programme under grant agreement No 955356.}
\thanks{$^{1}$ Are with the Autonomous Systems Lab, ETH Zurich, Zurich 8092, Switzerland, {\tt \small \{berlukas, lioott, rsiegwart, cesarc\}@ethz.ch.}}%
\thanks{$^{2}$ Are with the Robotics Systems Lab, ETH Zurich, Zurich 8092, Switzerland, {\tt \small \{skhattak, mhutter\}@ethz.ch.}}%
}
\begin{acronym}
\acro{SLAM}{Simultaneous Localization And Mapping}
\acro{TSDF}{Truncated Signed Distance Field}
\acro{ICP}{Iterative Closest Point}
\acro{RFS}{Random Finite Sets}
\acro{CNN}{Convolutional Neural Network}
\acro{CDF}{Cumulative Distribution Function}
\end{acronym}
\begin{document}
\maketitle
\thispagestyle{empty}
\pagestyle{empty}
\begin{abstract}

In this paper, we deal with the problem of creating globally consistent pose graphs in a centralized multi-robot SLAM framework. 
For each robot to act autonomously, individual onboard pose estimates and maps are maintained, which are then communicated to a central server to build an optimized global map.
However, inconsistencies between onboard and server estimates can occur due to onboard odometry drift or failure. 
Furthermore, robots do not benefit from the collaborative map if the server provides no feedback in a computationally tractable and bandwidth-efficient manner.
Motivated by this challenge, this paper proposes a novel collaborative mapping framework to enable accurate global mapping among robots and server. 
In particular, structural differences between robot and server graphs are exploited at different spatial scales using graph spectral analysis to generate necessary constraints for the individual robot pose graphs.
The proposed approach is thoroughly analyzed and validated using several real-world multi-robot field deployments where we show improvements of the onboard system up to 90\%.

\end{abstract}
\section{Introduction}
\label{sec:introduction}
Collaborative multi-robot exploration with heterogeneous platforms and sensors imposes significant challenges on current state-of-the-art mapping and localization approaches. 
Maintaining a consistent pose estimate across all employed systems is particularly difficult for distributed modules and is mission-critical for the operation of robot teams in applications like disaster response and search and rescue. 

With the recent availability of high-bandwidth mobile networks, e.g., 5G networks, centralized and collaborative robotic approaches have received increased attention in the robotics community due to their improved practical feasibility.
Typically, mobile robotic systems have limited onboard computational resources for which centralized approaches can assist by offloading computational intensive operations from individual robots to a shared centralized server.
A centralized server with more computational capacity can perform expensive operations such as global optimizations, loop closing, exploitation of all available sensor data to improve accuracy and also overcome onboard failure.
\begin{figure}[!t]
  \centering
   \includegraphics[width=0.5\textwidth, trim={0.1cm, 1.0cm, 0.2cm, 0cm}, clip]{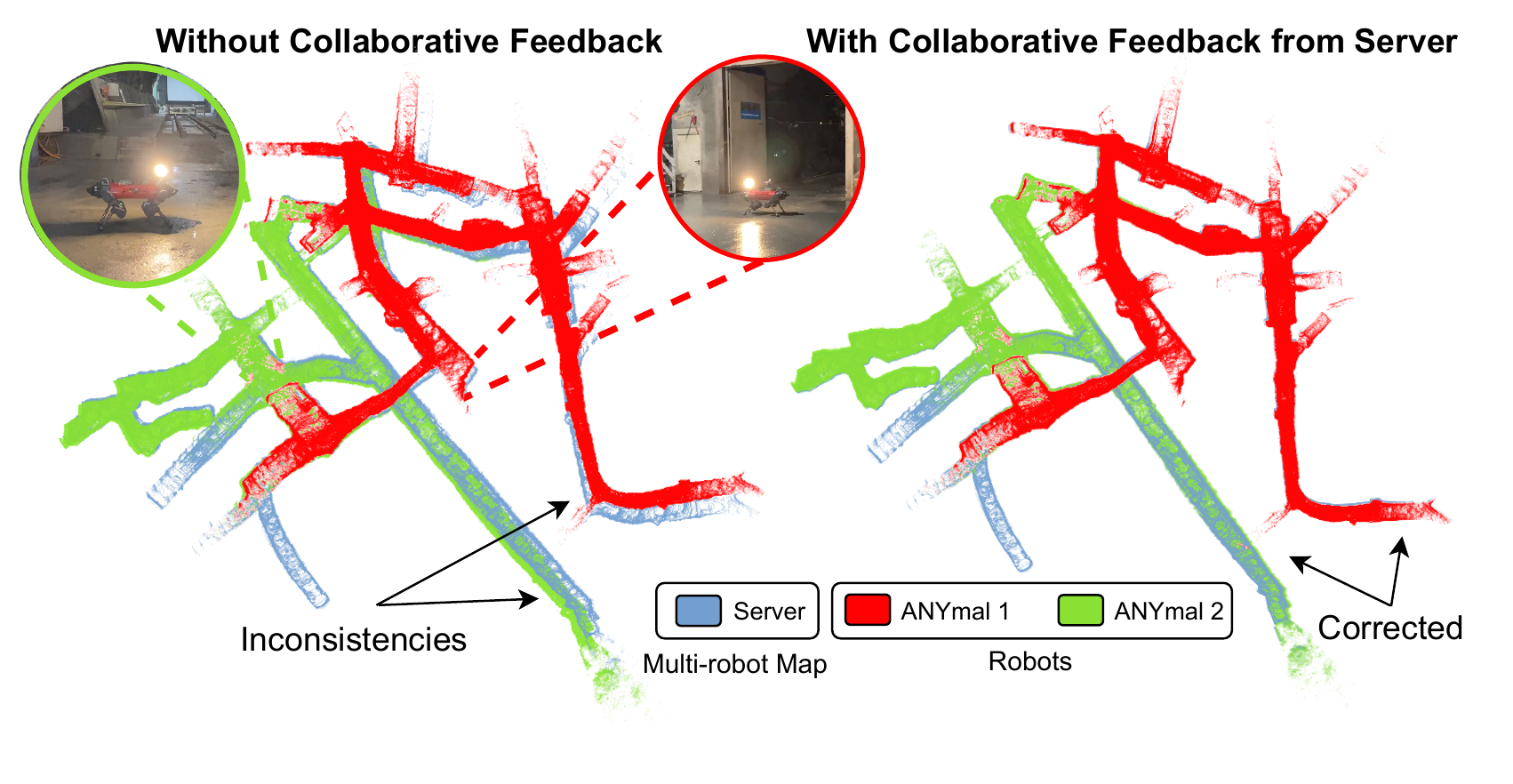}
   \caption{Overview of a large-scale multi-robot deployment in an underground tunnel system. Structural differences between single-robot maps and a collaborative global map (left) are used to derive necessary constraints for making the onboard and global estimates consistent (right).}
   \label{pics:introduction:hagerbach}
\end{figure}

%
Most collaborative mapping approaches focus on building accurate maps on the server and ignore the use of global multi-robot information to provide localization feedback to individual agents to make the onboard and server maps consistent w.r.t. each other.
Especially in centralized settings without feedback,
pose estimation discrepancies may arise between robots during large missions leading to severe drift between robot and server maps resulting in increased optimization time at the server for collaborative mapping. 
Therefore, it is evident that the provision of additional constraints to improve onboard estimation and collaborative mapping performance can be highly beneficial for large-scale multi-robot missions.

Furthermore, multi-robot missions often deploy a heterogeneous set of robots, e.g., aerial and ground robots, which depending on the type and role of the robot, carry a diverse set of multi-modal sensors onboard and correspondingly utilize different algorithms for robotic operations, e.g., localization and mapping.
Consequently, no common layer sharing data to improve pose estimation and mapping accuracy among the employed systems is readily available. Hence, a sensor modality-invariant approach that can incorporate and communicate relevant correction information among robots while maintaining low network bandwidth requirements is essential for large-scale multi-robot field deployments.
%
\begin{figure*}[!t]
  \centering
   \includegraphics[width=0.9\textwidth, trim={0.0cm, 0.0cm, 0.0cm, 0cm}, clip]{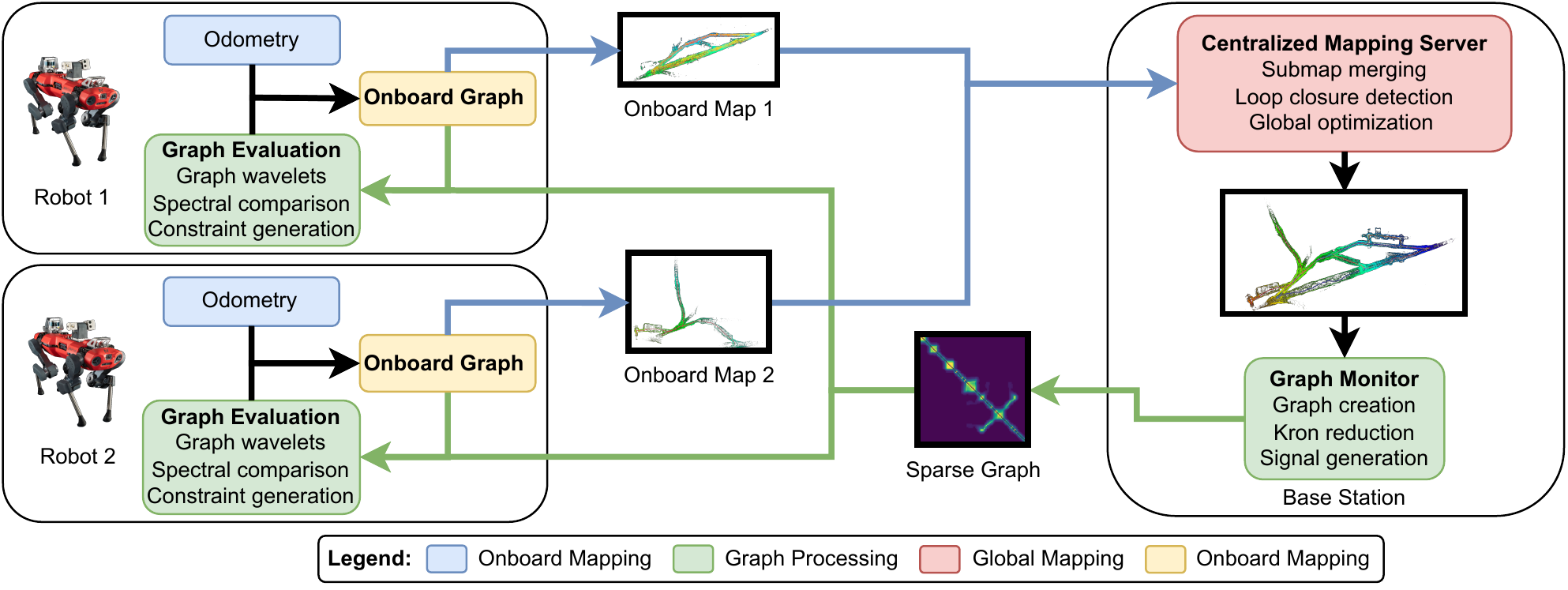}
   \caption{Our approach considers multiple robots individually exploring an environment and sending incremental mapping data to a centralized server that accumulates and jointly optimizes them. A relaxation of the collaborative multi-robot map is sent back to the robots, where a multi-scale graph spectral analysis is performed to identify discrepancies onboard and server maps and to generate necessary constraints for making them consistent.}
   \label{pics:introduction:summary}
\end{figure*}

Motivated by the discussion above, this paper proposes a novel collaborative multi-robot pose graph approach (cf. Figure~\ref{pics:introduction:summary}) which is independent of the underlying robot pose estimation processes and relies only on a sparse abstraction of the estimated poses -- a positional graph. 
Our framework operates in the graph spectral domain of the positional graphs to identify structural anomalies in the individual robot pose graphs using a multi-scale analysis.
By examining the structural components of the positional graphs at different scales, our system identifies discrepancies in the local and coarser neighborhoods and adds corresponding constraints to improve pose estimation accuracy of individual robots and to make the individual robot and collaborative server maps consistent.
Key contributions of this paper are:
\begin{itemize}
    \item Graph spectral analysis of positional graphs to identify disagreements between onboard and server pose graphs.
    \item Automatic adaptive inference of multi-scale constraints to achieve a consensus  between robot and server maps.
    \item Comparison against current state-of-the-art approaches on datasets and a thorough quantitative analysis on large-scale multi-robot field deployments are presented to validate the proposed approach.
\end{itemize}

\section{Related Work}\label{sec:related_work}
In this section, we review the state-of-the-art in collaborative and consistent multi-robot localization and mapping approaches as well as the current applications of graph signal processing. 
\subsection{Globally Consistent Multi-Robot Mapping}
Collaborative multi-robot approaches can be distinguished into centralized~\cite{Deutsch2016a, Schmuck2019, Karrer2018} and distributed solutions~\cite{Cunningham2013, Dong2015}.
Deutsch et al.~\cite{Deutsch2016a} proposed a vision-based centralized multi-robot SLAM approach where a mapping server performs loop closures and replaces robot pose graphs with corrected graphs. A similar approach was proposed by~\cite{Schmuck2019} in which robots send local maps to a mapping server which then returns optimized keyframes and landmarks to each robot to include in their onboard optimizations. 
To improve the speed of onboard optimizing tasks, CoSLAM~\cite{Zou2013} proposes to make use of GPU computing, hence requiring a GPU onboard individual robots. 
Different from vision-only approaches, the work of Ebadi~\cite{Ebadi2020} proposes a large-scale collaborative multi-modal SLAM framework.
However, their proposed approach does not provide any pose corrections from the centralized server to the individual robots.

In contrast to collaborative approaches, distributed approaches require each robot to run a full onboard SLAM solution~\cite{Dong2015} and share marginalized information with other robots~\cite{Cunningham2013}, thus making full information available to each robot but increasing the onboard compute requirements significantly.
Independent of the sensing modality used,~\cite{BeenKim2010} aims to achieve consistent maps across multiple robots by detecting loop closures between robots and connecting their pose graphs.
In this direction,~\cite{Mangelson2018a, Mangelson2019} aim to robustly select inter-robot loop closure candidates by maintaining pair-wise consistent measurements. 
More recently~\cite{Lajoie2020} proposed a distributed system with distributed loop closure detection. 

Most current collaborative multi-robot mapping approaches require the partial or complete substitution of the onboard pose graphs with the server-optimized graph. 
Conversely, this paper proposes to detect discrepancies between the robot graphs using spectral analysis and a sparse abstraction of the server graph to generate an individual set of constraints for each robot.
Hence, the proposed approach achieves high mapping accuracy  while maintaining low network and compute requirements.

\subsection{Graph Signal Processing}
Spectral graph theory is an active research area and has gained popularity in the past years in the context of robotics. Spectral graph theory approaches have been proposed for robotic mapping~\cite{Brunskill2007}, planning~\cite{Indelman2018}, and more recently, in combination with graph neural networks for various robotic tasks~\cite{Chandra2020, Moon2020}. 
In general, graphs are irregular structures and are capable of modeling large, complex, and distributed problems~\cite{Mateos2018}, e.g.~\cite{Egilmez2014} proposes an anomaly detection for spatial proximity of graph nodes using spectral graph filtering.
Furthermore, graph signal processing aims at applying signal processing techniques on graph structures, thus allowing the use of existing concepts such as the Laplacian operator~\cite{Sandryhaila2014} and multi-scale analysis~\cite{Hammond2011, Hammond2019}.
Similarly,~\cite{Donnat2018} aims to learn a multi-scale structural embedding using graph wavelets by treating the wavelet coefficients as a probability distribution. A good introduction and overview of graph signal processing is presented in~\cite{Ortega2017}. 

Our approach also performs a structural analysis of graph signals to detect discrepancies between the onboard and server graphs.
Using localized graph wavelets in the graph domain, our approach directly compares the trajectories at different scales to estimate the severity of the inconsistency of the individual positional graphs.

\section{Consistent Collaborative Multi-Robot Mapping}\label{sec:method}
This section details the proposed method, for which an overview is presented in Figure~\ref{pics:introduction:summary}.
Overall, the aim is to identify graphs nodes with high positional drift that will lead to large errors and correct them with only a few constraints. 
The proposed approach comprises of the following core components: (i) Onboard localization and mapping, (ii) Mapping server at the base station, (iii) Pose graph comparison and correction. 
\begin{figure}[!t]
  \centering
   \includegraphics[width=0.46\textwidth, trim={0.0cm, 0.1cm, 0.0cm, 0cm}, clip]{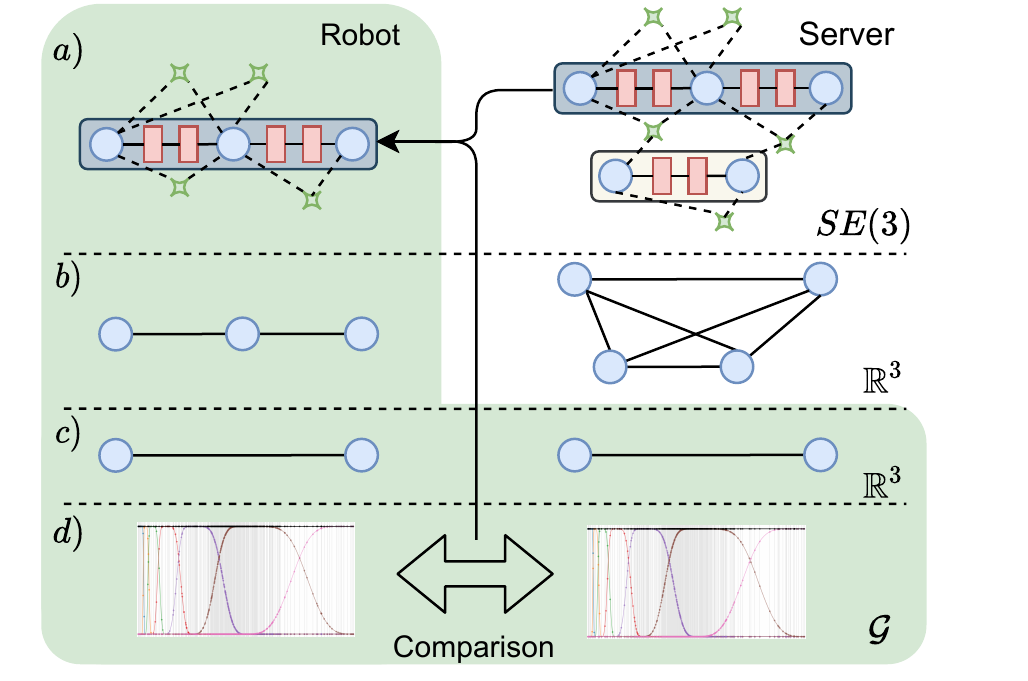}
   \caption{Different employed graphs: a) Onboard visual-inertial graph, which will be incorporated into the global server map. b) Positional graph proxies including Kron reduction. c) Synchronized graph proxies. d) Multiscale spectral graph analysis. With the results from (d) the system queries poses in $SE(3)$ to generate constraints.}
   \label{pics:method:graphs}
\end{figure}
\subsection{Onboard Localization and Mapping}\label{subsec:onboard}
Each robot performs onboard mapping and localization to provide an odometry estimate of its current position. 
This state estimate is then utilized to build a dense visual-inertial (VI) graph used for global multi-robot mapping on the server.
In brief, the odometry estimates are used to triangulate tracked visual features, and throughout the exploration, a factor graph in the form of keyframed~\cite{Dymczyk2015a} submaps is sent to the mapping server, where all submaps are accumulated, merged into a collaborative map, loop closed and globally optimized.
Consequently, all the computational-intense multi-robot operations are delegated to the central mapping server while the robots only perform the initial graph building.
\subsection{Centralized Mapping Server}
Each submap overlaps with the previous one, allowing the mapping server to readily combine submaps by attaching them, only requiring the handling of potential landmark conflicts.  
The mapping server holds and continuously operates on a single global multi-robot map. 
In particular, each iteration includes detection of for intra- and inter-robot loop closures~\cite{Lynen2015a} as well as performing a joint optimization on the merged multi-robot map.
Once a globally optimized map is available, the mapping server creates, relaxes and broadcasts the global graph.
Since for comparison rotational information is not required, the VI graph is relaxed to proxy graphs in $\mathbb{R}^3$ which contains only the positional information.

\textbf{Global Proxy Graph.}
The global proxy graph encapsulates the global knowledge of the environment in a compact representation and contains crucial information such as the last known positions of all robots. 
The global proxy graph is built by defining representative nodes for each incoming submap.
The representative nodes can be freely chosen but ought to reflect the robot trajectories to some degree. 
Moreover, since the global map is continuously optimized, the proxy graph is not immediately built but only a reference to each node is maintained.
When an update is triggered after the global optimization, the \textit{graph monitor} (cf. Figure~\ref{pics:introduction:summary}) retrieves the latest estimate of the multi-robot pose graph, builds the proxy graph, and sends it to all robots. 
Although only the proxy graph in $\mathbb{R}^3$ is needed for comparison, auxiliary information is also included for each node in the transmission, such as timestamps and orientations for synchronization and constraint generation purposes.

In general, the proxy graph is a weighted undirected graph $\mathcal{G}_{server}=(\mathcal{E},\mathcal{V},w)$ consisting of a set of nodes $\mathcal{V}$, edges $\mathcal{E}$ and weights $w: \mathcal{E}\mapsto\mathbb{R}^{+}$ denoting how strong two nodes are connected.
A radius search ($7\,\si{\meter}$) is performed around each vertex in the proxy graph and the weight $w_D$ of adjacent nodes is calculated using a squared exponential function where the weight decreases with increasing spatial distance, i.e.
\begin{equation}\label{eq:distance_weight}
    w_D(n,m)=\exp\left(-\frac{\norm{\mathcal{V}_n-\mathcal{V}_m}_2}{2\sigma^2}\right),
\end{equation}
with $\sigma$ being a free parameter.
The weighted adjacency matrix $\mat{A}$ is then given by $\mat{A}_{nm}=w_D(n,m)$. 

Consequently, the graph is described by $N$ nodes $n\in\mathbb{R}^3$ and its adjacency matrix $\mathbf{A}\in\mathbb{R}^{N\times{}N}$ with $\mathbf{A}_{nm} > 0$ if two nodes $n$ and $m$ are connected.
Furthermore, the degree matrix $\mathbf{D}$ is a diagonal matrix with entries $\mathbf{D}_{nn}=\sum_m \mathbf{A}_{nm}$ where $m$ are all incident nodes of $n$.
Based on these two matrices $\mathbf{A}$ and $\mathbf{D}$, the graph Laplacian $\mathcal{L}$ can be obtained by 
\begin{equation}\label{eq:laplace}
    \mathcal{L}=\mat{D}-\mat{A},
\end{equation}
which is by definition a symmetric and a positive semidefinite matrix.
Hence, it can be decomposed into its eigenvalues and eigenvectors, i.e. $\mathcal{L}=\mat{U}\Lambda{}\mat{U}^\top$.

As a final step, $\mathcal{G}_{server}$ is reduced using a Kron reduction~\cite{Dorfler2010}, when the graph increases to a specific number of nodes, to avoid long computation times.
The Kron reduction takes the nodes to keep from the proxy graph as input and reduces them while preserving the spectral properties and the adjacency matrix. 
The reduced nodes are selected based on the polarity of the largest eigenvector $u_{\text{max}}$ in $\mathcal{L}$~\cite{Shuman2016}, i.e.
\begin{equation}
    \mathcal{V}_{\text{reduced}} = \left\{n\in\mathcal{V}:u_{\text{max}}(n)\geq0\right\}.
\end{equation}
Finally, the global proxy graph $\mathcal{G}_{server}$ is broadcasted over the network. Each robot then uses this information to build an onboard graph $\mathcal{G}_{robot}$ using the procedure described above and synchronizes the nodes based on their timestamp (cf. Figure~\ref{pics:method:graphs}c). 
Next, we perform a graph spectral analysis using $\mathcal{G}_{server}$ and $\mathcal{G}_{robot}$ to identify drifts in the onboard estimation.

\subsection{Spectral Analysis of Graph Signals}\label{sec:method:spectral}
The analysis of spectral components of band-limited signals is a well-established and widely used technique in engineering and research.
In this paper, we utilize the theory of spectral analysis and signal processing defined on graphs.
In contrast to the standard spectral analysis, graphs do not assume any underlying manifold.
Therefore, they are well-suited for many robotic applications where graph structures such as pose graphs play a significant role.

The traditional Fourier transform is the inner product of a signal $x$ with a harmonic oscillation i.e.
\begin{equation}
    X(\omega)=\langle{}x(t), \exp\left(j\omega{}t\right)\rangle_t=\int_{-\infty}^{\infty} x(t)\exp\left(-j\omega{}t\right) dt,
\end{equation}
where $X(\omega)$ is the spectrum of $x(t)$ and $\exp\left(-j\omega{}t\right)$ the eigenfunctions of the one-dimensional Laplacian.
Analogously, the Graph Fourier transform $F$ of a function $f$ is given by the expansion of $f$ with the eigenfunctions $u$ of the graph Laplacian $\mathcal{L}$ (cf. Eq.~\eqref{eq:laplace}), i.e.
\begin{equation}
    F(\lambda_l)=\langle{}f(n), u_l(n)\rangle{}_n=\sum_nf(n)u_l^*(n),
\end{equation}
where $\lambda_l$ is the $l$-th non-negative eigenvalue of $\mathcal{L}$ corresponding to $u_l$ and $(\cdot)^*$ denotes a complex conjugation.

The graph Fourier transform of a signal $f$ is then $F = U^\top f$.
The eigenvalues $\Lambda$ are real values, thus can be ordered and correspond to the graph frequencies, allowing a similar intuition as for traditional frequency analysis.
Consequently, most of a graph signals energy is preserved in the lower bands of $\Lambda$ and higher bands correspond to high oscillating frequencies.

\textbf{Graph Comparison.}
After the robot has received a global update, a chronological synchronization is performed, yielding a one-to-one mapping of the global graph $\mathcal{G}_{server}$ and onboard graph $\mathcal{G}_{robot}$.
Next, the pose information at the nodes of each graph is used to create the functions $f$ and $h$ for server and robot, respectively. 
In particular, the positional information for each node in the graph is used to compute the relative distance to the origin.
Since rotational drifts result in positional structural discrepancies, the inclusion of rotational information is omitted and left for future research. 

Generally, wavelets are well-known to be very efficient and flexible for a variety of different tasks in signal processing problems~\cite{Hammond2019}. 
In traditional wavelet analysis, a signal $x(t)$ is projected onto a scaled ($a$) and shifted ($b$) wavelet $\psi$, i.e.
\begin{equation}\label{eq:classical_wt}
    W(a,b)=\langle{}x(t), \psi_{a,b}(t)\rangle_t=\int_{-\infty}^{\infty} \frac{1}{a}\psi^{*}\left(\frac{t-b}{a}\right)x(t)\,dt.
\end{equation}
Using Parseval's theorem, Eq.~\eqref{eq:classical_wt} can also be expressed with the Fourier-transformed signal: $W(a,b)=\langle{}X(\omega),\Psi_{a,b}(\omega)\rangle{}_\omega$.
Analogously, the graph wavelet transform can be derived using the graph Fourier transform and a wavelet filter kernel on $\mathcal{L}$.
For more details, we refer the interested reader to the work of Hammond et. al~\cite{Hammond2011, Hammond2019}.

In this work, the Meyer wavelet using seven scales ($s_{\text{max}}=7$) is used due to its good localization in the graph and frequency domain.
By construction, graph wavelets have the property of being localized on the graph and, therefore, directly relate to its structural properties.
The realization of a graph wavelet $\psi_{s,n}$ for a scale $s$ and node $n$ is given by 
\begin{equation}
    \mat{\psi}_{s,n} = \mat{U}\mat{G_s}(\Lambda)\mat{U}^\top\delta_n,
\end{equation}
where $\delta_n$ is a Dirac centered at vertex $n$ and $\mat{G_s}$ the wavelet filter bank at scale $s$.
In other words, the filter bank $\mat{G_s}$ acts only on the eigenvalues of the graph, i.e. $\mat{G_s}(\Lambda)=\text{diag}(g(s\lambda_1), \ldots, g(s\lambda_N)$ and is multiplied with the graph Fourier-transformed Dirac, followed by an inverse transform $\mathbf{U}$.
Since $\mat{\psi}_{s,n}$ lies in the graph domain, we can compute the wavelet coefficients for a graph signal $f$ using $\mat{W}_{s,n}=\mat{\psi}_{s,n}^\top{}f$.

In summary, the wavelet coefficients up to scale $s_{\text{max}}$ constitute a feature vector that represents multiscale structural information for a node $n$.

\subsection{Correcting Onboard Estimation}\label{sec:method:correction}
For a server node $n$ with corresponding onboard node $n'$, the scale-wise distance is computed as:
\begin{equation}
    d_{n,n'}^s = \norm{\mat{W}_{s,n} - \mat{W}_{s,n'}}_2,
\end{equation}
where $\mat{W}_{s,n}$ and $\mat{W}_{s,n'}$ were computed using $\mathcal{G}_{server}$ and $\mathcal{G}_{robot}$, respectively.

\begin{figure}[!t]
  \centering
   \includegraphics[width=0.4\textwidth, trim={0.0cm, 0.2cm, 0.0cm, 0cm}, clip]{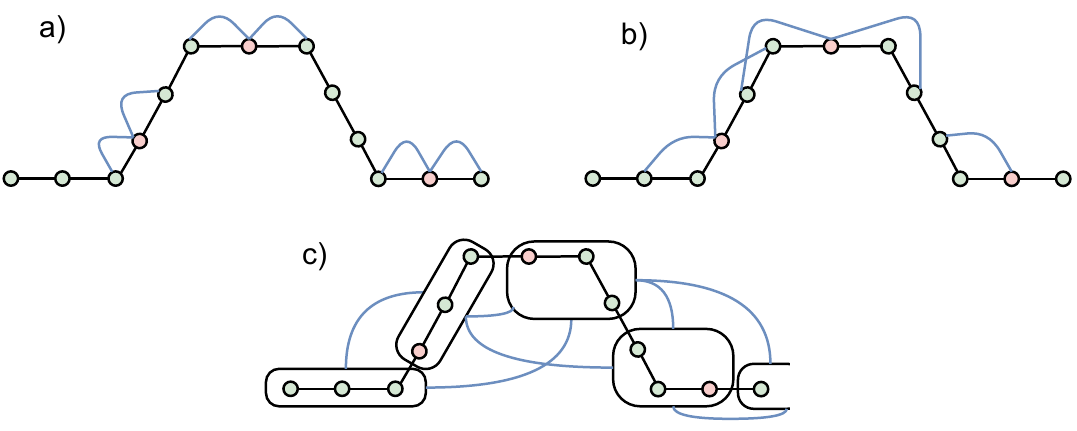}
   \caption{Illustration of three different relative constraint types. Based on the scale of the structural difference, additional constraints are added to correct (a) adjacent, (b) close neighborhood or (c) submaps.}
   \label{pics:method:constraints}
\end{figure}

Intuitively, since graph wavelets are localized at a specific node $n$ in the graph~\cite{Tremblay2014, Donnat2018}, large scales compress the filter function $g$ leading to a description of the larger neighborhood. 
In contrast, small scales stretch $g$ and thus yield a description of the closer neighborhood of $n$.
Consequently, three separate cases are distinguished (cf. Figure~\ref{pics:method:constraints}), i.e., a large difference in the lower, mid and higher scales of the coefficients. 
If $d_{n,n'}^s$ exceeds the threshold for small scales, a relative constraint is added between the direct neighbors of $n'$. 
Likewise, for mid-scale differences, a corresponding constraint within a two-node distance of $n'$ is added.
In the case of large-scale discrepancies, a constraint between the $k$-nearest submaps is added.
Since all different types of constraints are relative between nodes, the server and robot graph can be expressed in arbitrary and unknown frames.
Additionally, it should be noted that the comparison is based on the proxy graphs, but constraints originate from the optimized poses in $SE(3)$ (cf. Figure~\ref{pics:method:graphs}d). 

Furthermore, the existence of already added constraints between the corresponding nodes is checked before the addition of every new constraint to the graph, and constraints are only updated when there is a reasonable difference in translation and rotation.
Otherwise, they remain unchanged.

\section{Experiments}
\label{sec:experiments}
We thoroughly evaluate the proposed framework and demonstrate its real-world application using aerial and legged robot datasets.
First, we validate our approach and compare its performance to the current state-of-the-art methods using the EuRoC~\cite{Burri2016} dataset sequences to simulate multi-robot deployments.
Next, we demonstrate the real-world performance of our framework during a multi-robot autonomous exploration and mapping mission conducted in an underground tunnel system using ANYmal~\cite{ANYmal} legged robots. Finally, the localization recovery for an individual agent in case of onboard localization failure is demonstrated during a multi-robot experiment conducted in an environment consisting of indoor and outdoor areas.
For all experiments, the root-mean-square of the absolute trajectory error, denoted as RMSE, is used as an evaluation metric.
Furthermore, in each experiment, the close neighborhood constraints use a 5-hop distance, and the submap-constraints are added between the four closest submaps. Each robot creates a new submap every $10\,\si{\second}$ and sends it to the mapping server.
%
\subsection{EuRoC Dataset: Validation and Comparison}
To validate the proposed approach and compare its performance against the current state-of-the-art collaborative mapping frameworks~\cite{Karrer2018, Schmuck2019, Campos2021a}, Machine Hall (MH) sequences from the EuRoC dataset are used to evaluate single- and multi-robot performance. For each sequence, ROVIO~\cite{Bloesch2017} is used to provide monocular visual-inertial odometry for individual aerial agents. First, we evaluate the localization performance by comparing the onboard robot estimates, the collaborative server estimate, and the proposed approach to the provided ground truth; in addition, we also compare with current state-of-the-art approaches with results presented in Table~\ref{tab:euroc}. It can be noted that despite larger individual onboard error, the proposed framework still attains the lowest collaborative error. 
Furthermore, correcting the onboard estimation using our multi-scale spectral approach, the lowest single-robot errors are also achieved, demonstrating the proposed approach's effectiveness to correct large onboard estimation errors.

\begin{figure}[!t]
  \centering
  \includegraphics[width=0.30\textwidth, trim={0.0cm, 0.0cm, 0.0cm, 0cm}, clip]{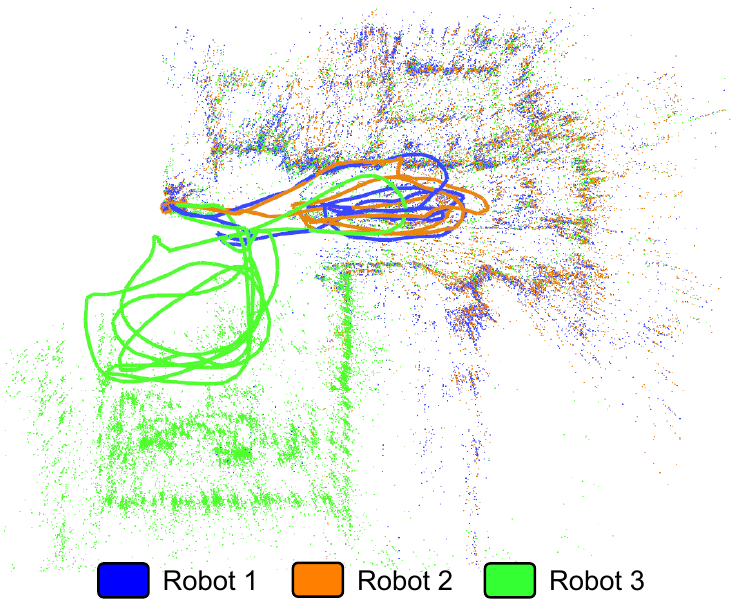}
  \caption{Collaboratvie VI multi-robot map built by three robots for the EuRoC dataset.}
  \label{pics:experiments:euroc}
\end{figure}

\begin{table}[!htb]
    \centering
    \begin{tabular}{cccc|c}
        \toprule
        \multicolumn{5}{c}{\textbf{EuRoC Machine Hall - Single and Collaborative}} \\
        \midrule
        \textbf{Method / Seq} & \textbf{MH01} & \textbf{MH02} & \textbf{MH03} & \textbf{MH01-03} \\
        VINS-mono\tablefootnote{Single robot results from~\cite{Qin2018}. Collaborative result from~\cite{Schmuck2019}.}~\cite{Qin2018} & 0.12\,\si{\metre} & 0.12\,\si{\metre} & 0.13\,\si{\metre} & 0.074\,\si{\metre} \\ 
        ORB-SLAM3\tablefootnote{Monocular visual-inertial results from~\cite{Campos2021a}.}~\cite{Campos2021a} & 0.062\,\si{\metre} & 0.037\,\si{\metre} & 0.046\,\si{\metre} & 0.037\,\si{\metre} \\ 
        Onboard  & 0.21\,\si{\metre} & 0.29\,\si{\metre} & 0.41\,\si{\metre} & \textbf{0.025}\,\si{\metre} \\ 
        \midrule
        CCM-SLAM\tablefootnote{As reported in~\cite{Schmuck2019}.}~\cite{Schmuck2019}  & 0.061\,\si{\metre} & 0.081\,\si{\metre} & 0.048\,\si{\metre} & 0.077\,\si{\metre} \\ 
        Proposed & \textbf{0.029}\,\si{\metre} & \textbf{0.028}\,\si{\metre} & \textbf{0.033}\,\si{\metre} & \textbf{0.025}\,\si{\metre} \\ 
        \bottomrule
    \end{tabular}
    \caption{RMSE comparison for the EuRoC dataset. Top part shows the results of single and collaborative approaches while the bottom row shows the individual corrected results.}
    \label{tab:euroc}
\end{table}

Next, using the experimental setup described in CVI-SLAM~\cite{Karrer2018}, it is demonstrated that the proposed approach can facilitate accurate pose estimation for individual robots by providing collaborative corrections, as shown in Table~\ref{tab:euroc:collab}.
\begin{table}[!htb]
    \centering
    \begin{tabular}{c|cccc}
        \toprule
        \multicolumn{5}{c}{\textbf{EuRoC Machine Hall - Collaborative Corrections}} \\
        \midrule
        \textbf{Sequences} & \multicolumn{2}{c}{\textbf{CVI-SLAM}~\cite{Karrer2018}} & \multicolumn{2}{c}{\textbf{Proposed}} \\
        & Single & Multi & Single & Multi \\
        MH01 \& MH02 & 0.224\,\si{\metre} & 0.139\,\si{\metre} & 0.29\,\si{\metre} & \textbf{0.030\,\si{\metre}} \\ 
        MH02 \& MH03 & 0.295\,\si{\metre} & 0.256\,\si{\metre} & 0.41\,\si{\metre} & \textbf{0.033\,\si{\metre}} \\
        MH04 \& MH05 & 0.412\,\si{\metre} & 0.34\,\si{\metre} & 0.62\,\si{\metre} & \textbf{0.094\,\si{\metre}} \\
        \bottomrule
    \end{tabular}
    \caption{Onboard pose RMSE after adding constraints from the centralized server for different dataset combinations.}
    \label{tab:euroc:collab}
\end{table}

\begin{figure}[!htb]
  \centering
  \includegraphics[width=0.45\textwidth, trim={0.0cm, 0.12cm, 0.0cm, 0.1cm}, clip]{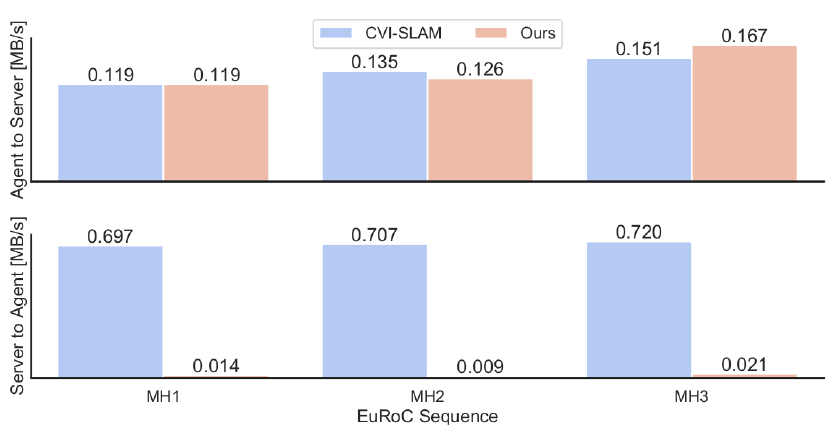}
  \caption{Average network usage for bi-directional communication between robot and server on the EuRoC dataset.}
  \label{pics:experiments:euroc_bw}
\end{figure}

Finally, to understand the potential for real-world applications, bi-directional network bandwidth utilization between robot and server is analyzed and compared to the network requirements of CVI-SLAM~\cite{Karrer2018}. The results are presented in Figure~\ref{pics:experiments:euroc_bw} and show that the robot-to-server network requirements of the proposed approach are comparable to CVI-SLAM.
However, the server-to-robot communication requirements are significantly reduced as only a sparse relaxation of the dense server graph is sent to the robots.

\subsection{Large-Scale Multi-Robot Subterranean Exploration} \label{exp:hagerbach}
To demonstrate the suitability towards complex real-world applications, the proposed approach was utilized during an autonomous multi-robot exploration~\cite{GBPlanner} and mapping mission conducted at Hagerbach underground facility in Switzerland.
Two ANYmal quadrupedal robots were deployed during an hour long mission and autonomously navigated distances of $1.21$\,\si{\km} and $1.1$\,\si{\km} respectively.
Each robot is equipped with a Velodyne VLP-16 LiDAR and a Sevensense Alphasense visual-inertial sensor. Onboard robot odometry estimation and mapping is performed by compSLAM~\cite{Khattak2020} and, along with required visual and pointcloud data, sent to the mapping server whenever the robots are within the communication range.

A ground-truth map was created using a Leica RTC360 scanner to evaluate the proposed collaborative mapping framework's performance and quantify the effect of integration of collaborative corrections on individual robot pose accuracy. Ground-truth robot poses were then computed by registering individual robot pointclouds against the ground-truth map following the approach of~\cite{ramezani2020newer}. 
The collaborative server and individual robots maps are shown in Figure~\ref{pics:introduction:hagerbach}, with quantitative results presented in Table~\ref{tab:exp:hagerbach}, demonstrating that the collaborative mapping approach achieves a lower error than either of the individual robot estimates.
In addition, benefiting from the feedback constraints, the onboard estimated pose error is significantly reduced.
Furthermore, the proposed approach adds the lowest number of additional constraints than the baseline approach of adding a correction to each node of the robot graph or for all nodes corresponding to the proxy graph while achieving comparable accuracy improvement without significantly increasing computational cost.
Finally, the robot-server network usage statistics for the mission are presented in Table~\ref{tab:bandwidth_results}, showing overall low communication bandwidth requirements and demonstrating the feasibility of the proposed approach towards real-world applications.
Additionally, by reducing the graph using the Kron reduction, we can further decrease the bandwidth by more than 40\%. 
\begin{table}[!htb]
    \centering
    \begin{tabular}{cccc|c}
        \toprule
        \multicolumn{5}{c}{\textbf{Underground Tunnel - Ground Truth Evaluation}} \\
        \midrule
        \textbf{Method} & \textbf{RMSE} & \textbf{Time} & \textbf{\# Factors} & \textbf{Server} \\
        ANYmal 1 & 1.15\,\si{\metre} & 68.6\,\si{\ms}  & 29152 & \textbf{0.25\,\si{\metre}} \\
        Baseline & 0.51\,\si{\metre} & 187.2\,\si{\ms}  & 31736 (+2584) \\
        Proposed & \textbf{0.46\,\si{\metre}} & 71.2\,\si{\ms}  & 29898 (+746) &  \\
        \midrule
        ANYmal 2 & 0.97\,\si{\metre} & 53.7\,\si{\ms}  & 25435 & \textbf{0.59\,\si{\metre}}\\
        Baseline & 0.72\,\si{\metre} & 152.6\,\si{\ms}  & 27815 (+2380) \\
        Proposed & \textbf{0.67\,\si{\metre}} & 59.8\,\si{\ms}  & 26067 (+632) & \\
        \bottomrule
    \end{tabular}
    \caption{RMSE comparison for the original onboard, server and corrected onboard graph. The numbers in parenthesis denote the improvement and total amount of additional constraints, respectively.}
    \label{tab:exp:hagerbach}
\end{table}
%

\begin{table}[!htb]
    \centering
    \begin{tabular}{ccc|cc}
        \toprule
        \multicolumn{5}{c}{\textbf{Underground Tunnel - Average Bandwidth}} \\
        \midrule
        \multicolumn{3}{c}{Without Reduction} & \multicolumn{2}{c}{With Reduction} \\
        \textbf{Robot} & \textbf{Rob$\rightarrow$Srv} & \textbf{Srv$\rightarrow$Rob} & \textbf{Nodes} & \textbf{Srv$\rightarrow$Rob}\\
        ANYmal 1 & 0.32\,\si{\MBps} & 0.044\,\si{\MBps} & 
        1295$\rightarrow$691 & 0.025\,\si{\MBps}\\
        ANYmal 2 & 0.30\,\si{\MBps} & 0.044\,\si{\MBps} & 
        1191$\rightarrow$444 & 0.025\,\si{\MBps}\\
        \bottomrule
    \end{tabular}
    \caption{Average bandwidth usage for the multi-robot underground mission. Our approach maintains low bandwidth requirements with the server-to-robot communication being identical as the same global graph is transmitted to all robots.}
    \label{tab:bandwidth_results}
\end{table}
\begin{table}[!htb]
    \centering
    \begin{tabular}{cccc|c}
        \toprule
        \multicolumn{5}{c}{\textbf{Indoor/Outdoor Dataset - Ground Truth Evaluation}} \\
        \midrule
        \textbf{Method} & \textbf{RMSE} & \textbf{Time} & \textbf{\# Factors} & \textbf{Server} \\
        ANYmal 1 & 2.22\,\si{\metre} & 5.2\,\si{\ms} & 3317 & \textbf{0.21\,\si{\metre}} \\
        Proposed & \textbf{0.30\,\si{\metre}} & 5.9\,\si{\ms} & 3339 (+22) &  \\
        \midrule
        ANYmal 2 & 0.25\,\si{\metre}  & 3.4\,\si{\ms}  & 2239 & \textbf{0.14\,\si{\metre}} \\
        Proposed & \textbf{0.16\,\si{\metre}} & 4.1\,\si{\ms}  & 2253 (+14) & \\
        \bottomrule
    \end{tabular}
    \caption{Comparison of the RMSE of the onboard estimation before and after the supplying additional constraints.}
    \label{tab:exp:localization_failure}
\end{table}
\begin{figure}[!htb]
  \centering
  \includegraphics[width=0.45\textwidth, trim={0.5cm, 0.3cm, 1.5cm, 0.1cm}, clip]{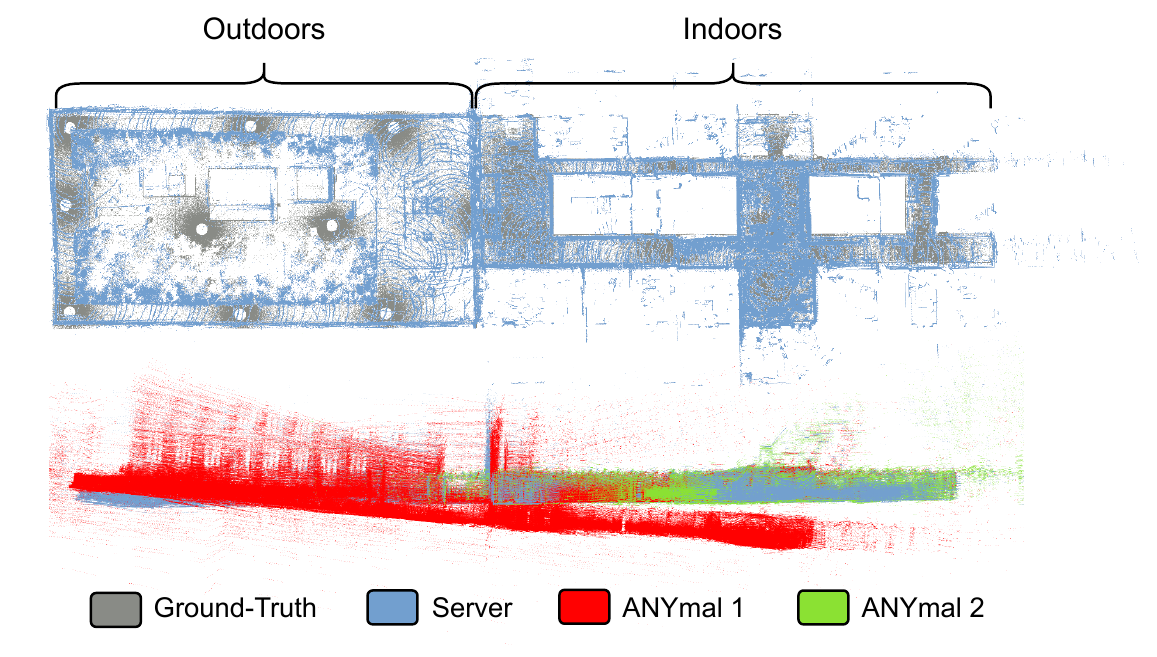}
  \caption{Mapping results for the indoor/outdoor multi-robot experiment. The top compares the collaborative map to the ground-truth while the bottom row shows a sideview comparison of the robot and server maps. The robot map of ANYmal 1 is misaligned due to onboard localization failure.}
  \label{pics:experiments:terrace}
\end{figure}
\subsection{Localization Recovery}
We conducted an experiment in a particularly challenging environment for LiDAR localization due to the absence of surrounding geometric structure to demonstrate the utility of collaborative mapping towards localization recovery for an individual robot in case of an onboard estimation failure.
Two ANYmal robots, equipped with the same sensory payload as described in subsection~\ref{exp:hagerbach}, were simultaneously deployed in an indoor office environment connected to an outdoor rooftop terrace. The first robot performs a loop indoors while the second robot transitions outdoors through a narrow doorway, navigates a rectangular path, and returns indoors. Due to the absence of surrounding structure outdoors, the onboard localization drifts significantly, skewing the onboard robot map. Nevertheless, the collaborative mapping approach is able to generate a consistent map of the environment, as shown in Figure~\ref{pics:experiments:terrace}, due to its loop closure capabilities. Furthermore, the integration of collaborative constraints enables localization recovery for the individual robot leading to a significant reduction in its pose error, as shown in Table~\ref{tab:exp:localization_failure}, when compared to ground-truth robot trajectory, generated as described in subsection~\ref{exp:hagerbach}.  

\section{Conclusions}\label{sec:conclusion}
This paper presented a novel framework for creating globally consistent estimates between multiple robots and a centralized mapping server. 
A graph-based spectral analysis of the robot and server graphs is proposed to identify the underlying structural differences in positional graphs.
By adding constraints on drifting nodes efficiently, low additional computation and communication resources are achieved.
The presented results using large-scale multi-robot field deployments in challenging environments demonstrate the real-world potential of the proposed approach for both research and industry.

%
%
\bibliographystyle{IEEEtran}
\bibliography{ref/references_cdpgo.bib}
\end{document}